\documentclass[final]{cvpr}

\usepackage{times}
\usepackage{epsfig}
\usepackage{graphicx}
\usepackage{amsmath}
\usepackage{amssymb}
\usepackage{booktabs}
\usepackage{multirow}

\usepackage[pagebackref=true,breaklinks=true,colorlinks,bookmarks=false]{hyperref}

\def\cvprPaperID{****} 
\def\confYear{CVPR 2021}

\begin{document}

\title{Winning the CVPR'2021 Kinetics-GEBD Challenge: \\
Contrastive Learning Approach}

\author{
    Hyolim Kang$^{1}$\\
    {\tt\small hyolimkang@yonsei.ac.kr}
    \and
    Jinwoo Kim$^{1}$\\
    {\tt\small jinwoo-kim@yonsei.ac.kr}
    \and
    Kyungmin Kim$^{1}$\\
    {\tt\small kyungminkim@yonsei.ac.kr}
    \and
    Taehyun Kim$^{1,2}$\\
    {\tt\small kimth0101@yonsei.ac.kr}
    \and
    Seon Joo Kim$^{1}$\\
    {\tt\small seonjookim@yonsei.ac.kr}
}

\maketitle

\begin{figure*}[t]
    \centering
    \includegraphics[width=\linewidth]{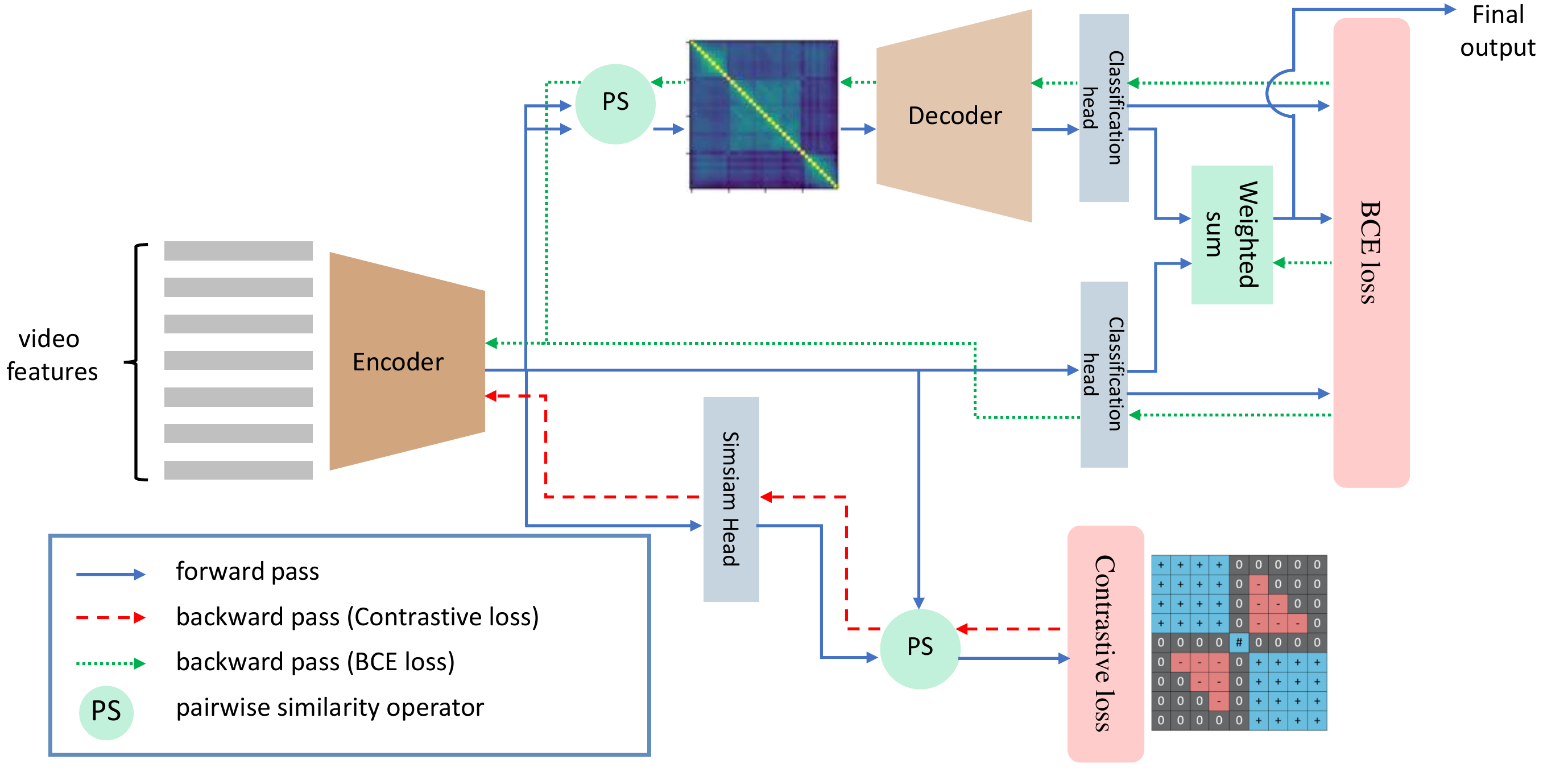}
    \caption{\textbf{Overall structure of our model.} Consecutive video frames are converted to video features (represented as gray bars in the figure) by a pre-trained network (omitted for brevity). As the green and red arrows show, the encoder receives gradient signals from both contrastive loss and BCE loss, allowing TSM to serve as a richer intermediate representation.}
    \label{fig:fig_model}
\end{figure*}
\begin{abstract}
    Generic Event Boundary Detection (GEBD) is a newly introduced task that aims to detect ``general'' event boundaries that correspond to natural human perception. In this paper, we introduce a novel contrastive learning based approach to deal with the GEBD. Our intuition is that the feature similarity of the video snippet would significantly vary near the event boundaries, while remaining relatively the same in the remaining part of the video. In our model, Temporal Self-similarity Matrix (TSM) is utilized as an intermediate representation which takes on a role as an information bottleneck. With our model, we achieved significant performance boost compared to the given baselines. Our code is available at \href{https://github.com/hello-jinwoo/LOVEU-CVPR2021}{\footnotesize\texttt{https://github.com/hello-jinwoo/LOVEU-CVPR2021}}.
\end{abstract}
\section{Introduction}
Cognitive science suggests when human perceives a long-term video, he or she  spontaneously parses the video in terms of ``event''~\cite{reisberg2013oxford}. 
To mimic this human visual perception, Generic Event Boundary Detection (GEBD)~\cite{shou2021generic}, which aims to detect general event boundaries that meet natural human criteria, is newly introduced.
Unlike previous tasks such as Temporal Action Localization (TAL)~\cite{shou2016temporal} or Shot Boundary Detection~\cite{baraldi2015shot}, GEBD's objective is to find out class-agnostic event boundaries, regardless of their categories.
As the groundtruth boundaries in GEBD are only defined by human's unknown perception process, the task is subjective in nature, which makes it more challenging compared to other temporal detection tasks~\cite{shou2016temporal,baraldi2015shot}.

To solve GEBD, we devised a novel network that uses temporal self-similarity matrix (TSM) as its intermediate representation.
The existence of an event boundary in a video implies that there is a visual content change at that point.
So, our main hypothesis is that similarity among the video snippet features would significantly vary near the event boundaries while relatively consistent in the other part of the video.
Accordingly, we may be able to infer where the action boundaries are by observing the TSM.

Assuming the hypothesis is true, a straightforward approach would be an neural network architecture similar to ~\cite{dwibedi2020counting}.
However, we found that BinaryCrossEntropy loss cannot provide sufficient gradients to form distinguishable TSM, especially when we adopt Transformers-based encoder architecture.
To compensate for insufficient gradient, we explicitly exploit the aforementioned hypothesis by employing the popular contrastive learning method~\cite{chen2020exploring} that showed promising performance in self-supervised learning.
With auxiliary contrastive loss, TSM can deliver more representative feature in terms of event boundary detection.

On the other hand, a direct approach without using TSM could provide a prediction with different perspective, which can improve overall model performance when combined with TSM-based prediction.
To this end, our final model (Figure~\ref{fig:fig_model}) has two passes: i) encoder-TSM-decoder-prediction (TSM pass) and ii) encoder-prediction (direct pass).
Both passes share the encoder part, and the convex combination of predictions from two passes is considered as a final output.

With our model and some training tricks, we achieved substantial performance boost compared to the given baselines in Kinetics-GEBD.

\section{Related Work}
\subsection{Temporal Self-Similarity Matirx (TSM)}
There have been notable works that used TSM to human action recognition~\cite{junejo2010view, korner2013temporal, sun2015exploring} as it can robustly express frame similarity regardless of large view point change if paired with an appropriate feature extractor.
Especially, ~\cite{dwibedi2020counting} proposed a novel TSM-based network architecture, RepNet, to solve action counting problem.
Note that in RepNet, the frame feature extractor is solely trained with the main task loss, while ours additionally utilized auxiliary contrastive loss term to reinforce weak gradient signal from the main (BinaryCrossEntropy) loss.
\subsection{Contrastive Learning}
In contrastive learning, positive samples are attracted together while the negative ones are repulsed. In spite of its frustratingly simple idea, it has shown clear performance gain in the self-supervised learning domain, resulting in seminal works including ~\cite{chen2020exploring, grill2020bootstrap, chen2020improved, chen2020simple}.
Among them, our model was particularly inspired by~\cite{chen2020exploring}, which adopted a simple siamese network with additional prediction head and ``StopGradient'' operator to conduct contrastive learning without negative samples.
Yet we borrowed the architectural concept from~\cite{chen2020exploring}, our model still exploits negative samples in the  contrastive loss term.
\begin{figure}[t]
    \centering
    \includegraphics[width=\linewidth]{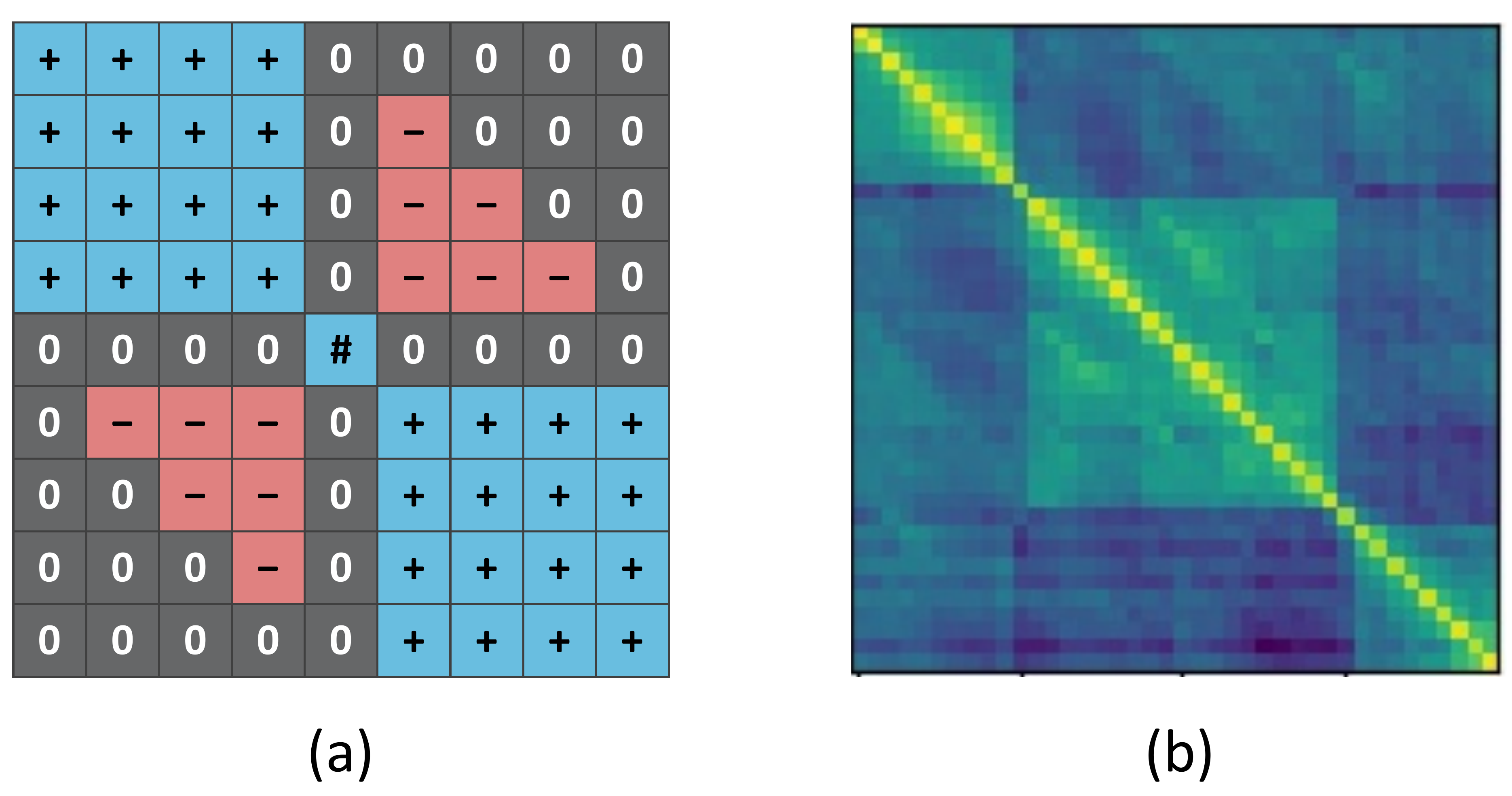}
    \caption{(a) explains how we define positive/negative/neutral sample for our contrastive loss (local range=4). `\#' sign at the center represents an event boundary, while blue`+'/red`-'/gray`0' cell denote positive/negative/neutral samples respectively. (b) is a sample of an actual pairwise similarity score matrix that is used to calculate the contrastive loss. A ternary mask that looks like (a) is applied to (b) to compute the contrastive loss. Note that (b) is slightly different from the TSM that goes into the decoder since it is pairwise similarities between the original encoded feature and encoded feature that has gone through the additional simsiam heads.}
    \label{fig:fig_mask}
\end{figure}
\section{Proposed Method} \label{Proposed Method}
\subsection{Encoder}
To capture various aspects of the given feature sequence, our encoder consists of a number of parallel modules each of which enjoys different receptive field.
For instance, 1D Convs with small/intermediate kernel size are used to capture local aspect of the feature sequence, while Transformer encoder architecture is employed to deal with long-term dependencies between the features.

Moreover, as Kinetics-GEBD dataset provides boundary classes, separate modules are allocated to detect each boundary class: action boundary, shot boundary, whole (action + shot) boundary.
It means that in the encoder, there are 3*4=12 parallel neural networks.

Furthermore, it would worth noting that the output features from each module are \textbf{NOT} concatenated until they are respectively converted to TSM (or pairwise similarity score matrix for the contrastive loss) form.
Only after the PS operation in Figure~\ref{fig:fig_model}, channelwise stacking operation of similarity matrices takes place, resulting in 12 channels in the TSM.

\subsection{Contrastive Loss}
For contrastive learning, we assume \emph{local} similarities between video features, unless there is an event boundary between them.
Concrete instantiation of the idea is given in Figure~\ref{fig:fig_mask}~(a).
Similarity scores between two \emph{distant} features (in Figure~\ref{fig:fig_mask}~(a), valid local range is set to 4), and between boundary element and the others are marked as neutral samples because of their ambiguity, while other \emph{local} scores are marked as positive or negative samples.

When we calculate the contrastive loss term, we just use positive/negative samples, ignoring neutral ones.
Let $i_k$ and $j_k$ denote a $k$th positive/negative sample respectively and $m$ and $n$ represent the number of positive/negative samples.
Then, the contrastive loss term $L_{contra}$ is defined as follows:
\begin{equation}
    L_{contra}=\frac{1}{m}\sum_{k=1}^mj_k-\frac{1}{n}\sum_{k=1}^ni_k
\end{equation}

\subsection{Other Modules}
To begin with, let $L$ stand for the feature length.
As a backbone architecture for the decoder, ResNet-18 architectue is slightly modified to take $(B,12,L,L)$ tensor as its input and output the $(B,C_{decoder},L,L)$ size tensor.
From $(B,C_{decoder},L,L)$ tensor, we gather diagonal elements, resulting in a tensor with shape  $(B,C_{decoder},L)$.
The boundary predictions are made with the shallow Conv1d classification head, which takes $(B,C_{decoder},L)$ tensor as its input.

On the other hand, we used Transformer encoder architecture to the classification head in the direct pass, since it should directly process relatively high-dimensional features compared to the classification head in TSM pass.
The Simsiam head is just two-layer 1D Convs with $kernel\_size=1$. A convex combination parameter in Figure~\ref{fig:fig_model}'s weighted sum operation is learnt by gradient from BCE loss.
Final loss term is a linear combination of three BCE loss and $L_{contra}$.

For more detailed architectural information, please refer our released code \href{https://github.com/hello-jinwoo/LOVEU-CVPR2021}{\footnotesize{https://github.com/hello-jinwoo/LOVEU-CVPR2021}}.

\section{Experiments}
\subsection{Dataset and Feature}
As the competition is targetting on the Kinetics-GEBD dataset, we only used the Kinetics-GEBD data for training.
No additional data or annotation is exploited during training procedure, regardless of track 1.1 and track 1.2.

Following the prevailing convention in Online Action Detection (OAD)~\cite{xu2019temporal, de2016online} and Temporal Action Localization (TAL)~\cite{shou2016temporal, shou2017cdc, zeng2019graph, xu2020g, yeung2016end, lin2020fast, bai2020boundary}, pre-trained feature extractor is applied to extract snippet-wise features.
For track 1.2, we used Kinetics pre-trained two-stream TSN~\cite{wang2016temporal} and SlowFast~\cite{feichtenhofer2019slowfast} features, while for track 1.1 we additionally used Activitynet~\cite{caba2015activitynet} pre-trained TSP~\cite{alwassel2020tsp} features.
For clarification, we will leave the source from which we downloaded the pre-trained weights on the footnote.\footnote{SlowFast: R50-8x8 from \href{https://github.com/facebookresearch/pytorchvideo/blob/master/docs/source/model_zoo.md}{pytorchvideo}}
\footnote{TSN: Inference code is based on \href{https://github.com/yjxiong/anet2016-cuhk}{anet2016-cuhk}~\cite{xiong2016cuhk} and the model weight is from \href{http://yjxiong.me/others/kinetics_action/}{here} (Inception V3)~\cite{wang2016temporal}.}
\footnote{TSP: \href{https://github.com/HumamAlwassel/TSP}{https://github.com/HumamAlwassel/TSP}}
\subsection{Ablation Study}
Table~\ref{tab:my-table} illustrates the result of ablation study for our model.
As shown in the table, each component of our model helps improve the performance when it is combined with each other.
Especially, there is a huge gap between our full model and the model without TSM pass (+3.3 in f1 score).
Besides, we found that TSM pass without CL (Contrastive Loss) does not work well, indicating that the sole BCE loss cannot deliver sufficient gradient to train the whole network.
We achieved our best performance model by incorporating both direct pass and TSM pass.

\begin{table}[]
\centering
\resizebox{\columnwidth}{!}{
\begin{tabular}{clc}
\toprule
& \multicolumn{1}{l}{Model} & \multicolumn{1}{c}{f1 score}\\
\midrule
\midrule
\multirow{1}{*}{Baseline} & PC~\cite{shou2021generic}     & 62.5            \\ \midrule
\multirow{4}{*}{Ours} & Direct pass     & 78.0 (+15.5)           \\
 & TSM pass w/o CL & 74.3 (+11.8)           \\
 & TSM pass w/ CL   & 81.0 (+18.5)          \\
 & \textbf{Direct pass + TSM pass w/ CL}    & \textbf{81.3 (+18.8)} \\ \bottomrule
\end{tabular}
}
\vspace{+1.0mm}
\caption{Comparison of our methods with baseline PC model and ablation study for each module in our model. CL stands for the  Contrastive Loss. Note that we conducted this experiment on one of the five validation splits according to Section~\ref{Multiple Dataset Splits}}
\label{tab:my-table}
\end{table}

\subsection{Miscellaneous}
In this section, we introduce several tricks to further ameliorate the final performance.

\textbf{Post-processing.}
To prevent duplicate predictions of adjecent frames, we use peak-estimation technique after the network estimation. We predicted a frame as a boundary frame if and only if its probability score is higher than the neighboring K (we set K as 1 or 2) ones. To prune some low-probability peak from the prediction, we set the threshold so that the frame having probability score lower than the threshold is not predicted as a boundary frame. 

\textbf{Multiple Dataset Splits.} \label{Multiple Dataset Splits}
We made new train/val splits with larger train set size for the better use of the dataset.
Specifically, we arbitrarily split \texttt{train+val} set into 5 equal-sized, non-overlapping groups.
Then, we form 5 sets of train/val split by taking one group as a validation split and the rest of them as a training split.
We train a model on each split, resulting in 5 models in total, and take the mean value of their output to obtain the final prediction.

\textbf{Model Ensembles.}
We ensembled some predictions to boost the performance of our model and prevent it from overfitting on validation sets. We used the predictions from all 5 folds, and a simple additional network consisting of two branch, one of which contains Transformer encoder and RNN decoder, and the other of which contains only CNN layers.

\section{Conclusion}
In this paper, we introduced a novel method to solve  newly-introduced GEBD task.
Our model accomplished compelling performance gain compared to the given baselines.
However due to lack of time, we could not conduct exhaustive ablation study.
For future work, we will do extensive experiments to strengthen our idea.

\section{Acknowledgement}
This work was supported by Hyundai Mobis Co., Ltd.(CTR210200164).

{\small
\bibliographystyle{ieee_fullname}
\bibliography{egbib}
}

\end{document}